\documentclass[journal]{IEEEtran}
\IEEEoverridecommandlockouts
\usepackage{cite}
\usepackage{amsmath,amssymb,amsfonts}
\usepackage{algorithmic}
\usepackage{graphicx}
\usepackage{array}
\usepackage{booktabs}
\usepackage{threeparttable}
\usepackage{textcomp}
\usepackage{xcolor}

\begin{document}
\title{Self-Organizing Map assisted Deep Autoencoding Gaussian Mixture Model for Intrusion Detection}

\author{Yang Chen, 
    	Nami Ashizawa,
    	Seanglidet Yean,
    	Chai Kiat Yeo,
    	Naoto Yanai
    	\thanks{This study is supported by Grant No. NTU M4082227 and Innovation Platform for Society 5.0 at MEXT.}
    	\thanks{Y. Chen, S. Yean, and C. K. Yeo are with the School of Computer Science and Engineering, Nanyang Technological University, Singapore 639798, Singapore. (e-mail: yang.chen@ntu.edu.sg)}
    	\thanks{N. Ashizawa and N. Yanai are with the Graduate School of Information Science and Technology, Osaka University, Osaka 565-0871, Japan.}}
\markboth{}
{Shell \MakeLowercase{\textit{et al.}}: Bare Demo of IEEEtran.cls for IEEE Journals}

\maketitle
\begin{abstract}
In the information age, a secure and stable network environment is essential and hence intrusion detection is critical for any networks. In this paper, we propose a \textit{self-organizing map assisted deep autoencoding Gaussian mixture model (SOM-DAGMM)} supplemented with well-preserved input space topology for more accurate network intrusion detection. The deep autoencoding Gaussian mixture model comprises a compression network and an estimation network which is able to perform unsupervised joint training. However, the code generated by the autoencoder is inept at preserving the topology of the input space, which is rooted in the bottleneck of the adopted deep structure. A self-organizing map has been introduced to construct SOM-DAGMM for addressing this issue. The superiority of the proposed SOM-DAGMM is empirically demonstrated with extensive experiments conducted upon two datasets. Experimental results show that SOM-DAGMM outperforms state-of-the-art DAGMM on all tests, and achieves up to 15.58\% improvement in F1 score and with better stability. 
\end{abstract}


\section{Introduction}

A secure and stable network environment is critical in the digital and networked world. Hence network intrusion detection is essential to monitor the network for malicious activities or policy violations. Related studies have designed approaches to recognize patterns in network data that do not conform to normal behaviors. These abnormal patterns are often referred to as anomalies, outliers etc. in the various domains. With the proliferation of artificial intelligence, network-based anomaly detection now use machine learning and deep learning to create a trusted behavior model so as to detect suspicious incoming activities~\cite{tsai2009intrusion}. \par

Before the era of deep learning, the studies on intrusion detection systems mainly focused on employing classical machine learning methods, e.g. support vector machines (SVM), random forests and k-nearest neighbor (KNN). Nowadays, deep learning offers competitive alternatives since more complex algorithms coupled with greater computational capacity can be obtained~\cite{zhang2019efficient}. Academic and industrial realms have thus witnessed the rapid development of deep anomaly detection~\cite{chalapathy2019deep}, e.g. self-taught learning based deep learning~\cite{javaid2016deep,chen2018autoencoder,bao2018restricted} and deep autoencoding Gaussian mixture model (DAGMM)~\cite{zong2018deep}. DAGMM achieves state-of-the-art performance since the model consists of a compression network and an estimation network and is capable of getting itself trained in an end-to-end manner. It is thus characterized as a form of unsupervised joint training. \par

{DAGMM partially overcomes previous models' incapability of preserving essential information in the low-dimensional space. Apart from the compression network for dimensionality reduction, density estimation which is at the core of DAGMM assumes that anomalies reside in the low probability density region~\cite{zong2018deep}. In other words, data points that are similar in the input space are supposed to be close to one another in the output space of the compression network. Thus, the DAGMM faces a dilemma of choosing between the low-dimensional requirement of GMM and the aforementioned topological structure preserving need. In the original version, the adopted autoencoder, playing the role of compression network, is only a basic architecture and can scarcely preserve the raw feature space topology well thereby greatly limiting the performance of DAGMM.}\par

{In recent machine learning studies, an increasing usage of topological features has been witnessed~\cite{hofer2019connectivity, guss2018characterizing}. However, the similar methodology using topology directly as a constraint for current deep learning approaches remains a challenge given the gap between the inherently discrete nature of these topological data analysis computations and the backpropagation process. This issue is only possible to be addressed under certain special circumstances~\cite{poulenard2018topological}. }\par

Motivated by the aforementioned, we propose a \textit{self-organizing map assisted deep autoencoding Gaussian mixture model (SOM-DAGMM)} for intrusion detection. The two primary contributions are presented in this paper. 

First, SOM-DAGMM is able to better preserve the input space topology in comparison with DAGMM by virtue of self-organizing map (SOM). Since DAGMM relies on joint training based on backpropagation which \textit{conflicts with the direct use of topological features}, this thus leaves us no other choice but to adopt a two-phase approach, i.e. \textit{plugging pre-trained SOM encoding} into the DAGMM. SOM is selected by virtue of its good capability of preserving topology.

Second, the superiority of SOM-DAGMM compared to the original DAGMM is empirically demonstrated with extensive experiments as well. 
To demonstrate this point of our model, testing of SOM-DAGMM on two security datasets is conducted. SOM-DAGMM \textit{outperforms DAGMM on all standard metrics with more stable performance}. 

The remainder of this paper is organized as follows. The literature review regarding \textit{Intrusion Detection} and \textit{Self-Organizing Map} is presented in Section~\ref{sec_2}.
The proposed self-organizing map assisted deep autoencoding Gaussian mixture model (SOM-DAGMM) is detailed in Section~\ref{sec_3}.
Section~\ref{sec_4} presents details of the experiments conducted on the proposed SOM-DAGMM. Section~\ref{sec_5} concludes this paper. \par

\section{Related Work}\label{sec_2}

In early literature, we briefly describe the research on intrusion detection and the use of self-organizing map. 

\subsection{Intrusion Detection}
As a significant application of anomaly detection in network security, intrusion detection, also called network anomaly detection, has been studied for decades~\cite{edgeworth1887xli}.\par

Intrusion detection systems are typically divided into two categories, namely, signature (rule)-based and anomaly-based. In this study, the discussed model belonging to the latter group leverages machine learning approaches to determine whether the deviation from the established normal usage patterns can be flagged as intrusions~\cite{tsai2009intrusion}. Many studies employing classical machine learning models e.g. support vector machines (SVM), self-organizing map (SOM), naive Bayes networks have been conducted. However, they suffer from poor performance such as low accuracy and high False Positive Rate (FPR). \par

Inspired by a great number of breakthroughs in the various applications that have been achieved through deep learning, researchers have significantly improved the model complexity to achieve substantial performance improvement by adopting deep models~\cite{javaid2016deep, chalapathy2019deep}. Deep Autoencoding Gaussian Mixture Model (DAGMM) has been recently proposed in~\cite{zong2018deep} and it produces good results without the need to label the training data. Our work is thus motivated by DAGMM. Besides, Chen et al. presented a federated deep autoencoding Gaussian mixture model (FDAGMM) to improve the disappointing performance of DAGMM caused by limited data amount~\cite{chen2019network}.\par

\subsection{Self-Organizing Map}\label{2b}
Self-Organizing Map (SOM) is an unsupervised learning algorithm that uses Artificial Neural Network (ANN) to discretize inputs from training samples to the two-dimensional feature layer. SOM emphasizes on two key features: mapping input dataset to the low-dimensional map and preserving the topology of the input. It is trained by initializing a random vector from the map grids and continuously activating other map units by selecting weights that are close to the input space. This algorithm uses Euclidean Distance to measure the closeness of every node to the input vectors and Best Matching Unit (BMU) technique to choose the closest weight as the winner. Thereafter, the neighbors of the winning weight get rewarded by getting their weights updated. Hence, over time, the number of neighbors of BMU decreases, resulting in a general form of distribution with Self-Organizing Map competitive learning method~\cite{kohonen1982self}. \par

For intrusion detection, SOM has made great contributions in anomaly detection. SOM is a perfect fit for anomaly detection tasks due to the nature of dataset representation, which either mainly originates from a single class or has difficulty in obtaining all of the failure representations. In other words, this method would rather focus on the domain description than estimating the probability density. 
For example, SOM has been proposed as a failure detection process to describe the normal system and identify the irregular behavior with compatibility measure (i.e. provides information on the regularity of the input-output mapping)~\cite{ypma1997novelty}. 
Labib and Vemuri creates a prototype system, NSOM, using SOM to classify regular and irregular network traffic on a real-time Ethernet network specifically for time-sensitive attacks (Denial of Service), that involve several successive packets and targets a host in a finite time limit~\cite{labib2002nsom}. SOM preserves the relationship between senders, receivers and protocols during the mapping. 
Similarly, Ramadas et al. model multiple SOMs to capture the characteristic patterns for each network service, e.g. web and email~\cite{ramadas2003detecting}. The intrusion or abnormal behavior, is detected by its distance exceeding the pre-set threshold, which is outside the normal cluster. 
\par

There are many variations of SOMs being applied for anomaly detection. The main advantages of SOM are using competitive learning as each data point competes for representation as well as neighborhood function to preserve the topological properties of the input space by using BMU. 
Therefore, in this paper, we use a pre-trained SOM model to produce a part of input representation for the compressed network which will be explained in Section~\ref{sec_3}.
\par

\section{Self-Organizing Map assisted Deep Autoencoding Gaussian Mixture Model}\label{sec_3}

\begin{figure*}[!ht]
	\centering
	\includegraphics[width=5.6in, height=2.95in]{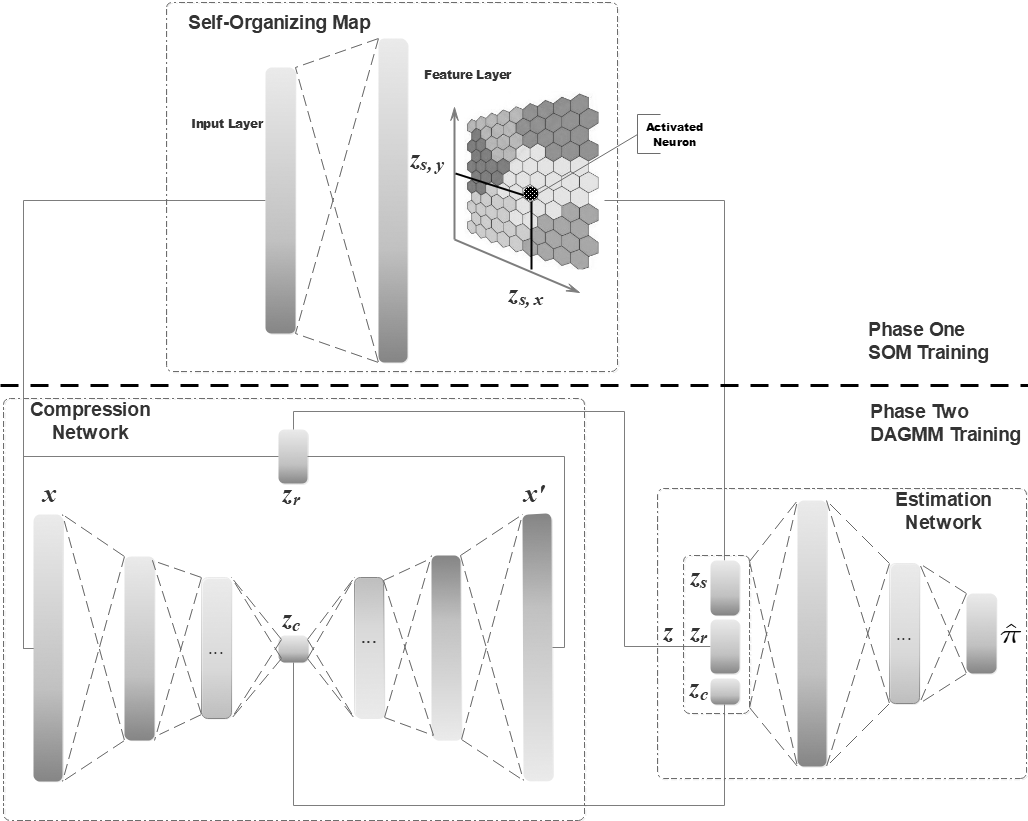}
	\caption{Overview of Self-Organizing Map assisted Deep Autoencoding Gaussian Mixture Model.}
	\label{Fig_overview_SOM_DAGMM}
\end{figure*}

As previously discussed, the issue of preserving the input space topology results in performance deterioration of DAGMM. 
Thus, the motivation of SOM-DAGMM is to address this issue through reserving the critical topological relationship well, which is one of SOM characteristics. 
In this section, we first describe the overview of SOM-DAGMM and then show details for each component of the model.\par

\subsection{Model Overview} \label{Overview}

The proposed model consists of three components, i.e. \textbf{Compression Network}, \textbf{Estimation Network}, and \textbf{SOM}. 
Fig.~\ref{Fig_overview_SOM_DAGMM} shows the overview of SOM-DAGMM. 
The lower components in the figure, i.e. the compression network and the estimation networks, are identical to DAGMM~\cite{zong2018deep} while the upper component is inspired by the self-organizing map~\cite{kohonen1982self}.

As per the main problem statement, DAGMM restricts the dimension of the autoencoder to one to suit the GMM model. Although a few more extra bits, e.g. the two bits corresponding to the two adopted distance metrics for measuring error construction, have been added, DAGMM can hardly preserve the input space topology well. As previously discussed, the issue is expected to be addressed with SOM.

{More specifically, due to some conflicts between the backpropagation that DAGMM relies on and the inherently discrete nature of topological data analysis computations, a two-phase training approach~\cite{candes2011robust} is adopted. In the first phase, the SOM training is first conducted. Taking the raw feature $ \mathbf{x} $ as input, the obtained model outputs SOM-generated low-dimensional representation $\mathbf{z}_{s}$.}

{In the second phase, DAGMM training is then carried out. Deep autoencoder takes $ \mathbf{x} $ as input to generate autoencoder-generated low-dimensional representation $ \mathbf{z}_{c} $ and reconstruction error $ \mathbf{z}_{r} $. Together with pre-generated SOM encoding $\mathbf{z}_{s}$, they are then concatenated to form $\mathbf{z}$. The low-dimensional representation can be redefined as:} 
\begin{equation}\label{eq_z}{
    \begin{aligned}
    \mathbf{z} &=\left[\mathbf{z}_{s}, \mathbf{z}_{r}, \mathbf{z}_{c} \right].\\
    \end{aligned}
}\end{equation}
{It is noted that in this second training phase, the SOM model and its corresponding encoding $\mathbf{z}_{s}$ remain fixed.}

\subsection{Self-Organizing Map}

As described in Section~\ref{2b}, SOM is based on ANN to discretize input from training samples to the two-dimensional feature layer. 
Similar to the dimensional reduction to~\eqref{eq_ae} described later,
the SOM-generated low-dimensional representation is defined as:
\begin{equation}\label{eq_zs}{
    \begin{aligned}
    \mathbf{z}_{s} &= som\left(\mathbf{x} ; \theta_{s}\right)\\
    \end{aligned}
}\end{equation}
where $som(\cdot)$ denotes the SOM processing. In particular, SOM takes $\mathbf{x}$ and returns the normalized coordinates of the activated neuron, i.e. $\mathbf{z}_{s} = (\mathbf{z}_{s, x}, \mathbf{z}_{s, y})$. Lastly, $\mathbf{z}$ is then fed to the subsequent estimation network to obtain the corresponding mixture membership prediction as described later. 
\par

\subsection{Compression Network}


With the raw feature of a sample denoted as $ \mathbf{x} $, the compression network that is implemented with a deep autoencoder conducts dimensionality reduction to output the corresponding low-dimensional representation $ \mathbf{z} $ as follows:
	\begin{equation}\label{eq_ae}{
		\begin{aligned}
		\mathbf{z}_{c} &=h\left(\mathbf{x} ; \theta_{e}\right),\\
		\mathbf{x}^{\prime} &=g\left(\mathbf{z}_{c} ; \theta_{d}\right),\\ 
		\mathbf{z}_{r} &=f\left(\mathbf{x}, \mathbf{x}^{\prime}\right),\\ 
		\mathbf{z}' &=\left[\mathbf{z}_{c}, \mathbf{z}_{r}\right],\\
		\end{aligned}
	}\end{equation}
where $ \theta_{e} $ and $ \theta_{d} $ correspond to the decoder and encoder parameters respectively, which are the two sub-components of the autoencoder. The encoder takes the raw feature $ \mathbf{x} $ and gives a low-dimensional representation $ \mathbf{z}_{c} $, while the decoder gets the representation as input and outputs $ \mathbf{x}^{\prime} $ as the reconstruction of $ \mathbf{x} $. $ \mathbf{z}_{r} $ indicates the reconstruction error. $ h(\cdot) $, $ g(\cdot) $, and $ f(\cdot) $ denote the encoding, decoding and reconstruction-error calculation function respectively.
As described above, by combining the above output with SOM, $\mathbf{z} =\left[\mathbf{z}_{s}, \mathbf{z}_{r}, \mathbf{z}_{c} \right]$ will be sent to the estimation network described below.

\subsection{Estimation Network}

 The estimation network gets $ \mathbf{z} $ from the compression network as its input. It conducts density estimation based on a Gaussian Mixture Model (GMM). A Multi-Layer Neural Network, denoted as $ M L N(\cdot) $, is utilized for predicting the mixture membership as follows:
	\begin{equation}\label{eq_GMM}{
		\begin{aligned}
		\mathbf{p} &=M L N\left(\mathbf{z} ; \theta_{m}\right), \\
		\hat{\gamma} &=\operatorname{softmax}(\mathbf{p}), \\
		\end{aligned}
	}\end{equation}
	where $ \theta_{m} $ stands for $ M L N $ parameters, $ K $ indicates the number of mixture components and $ \hat{\gamma} $ corresponds to a $ K $-dimensional vector predicting the soft mixture-component membership. With the batch size $ N $, GMM parameter estimation is performed as follows:
	\begin{equation}\label{eq_Est}{
		\begin{aligned}
		\hat{\phi}_{k} &=\sum_{i=1}^{N} \frac{\hat{\gamma}_{i k}}{N}, \\
		\hat{\mu}_{k} &=\frac{\sum_{i=1}^{N} \hat{\gamma}_{i k} \mathbf{z}_{i}}{\sum_{i=1}^{N} \hat{\gamma}_{i k}}, \\
		\hat{\mathbf{\Sigma}}_{k} &=\frac{\sum_{i=1}^{N} \hat{\gamma}_{i k}\left(\mathbf{z}_{i}-\hat{\mu}_{k}\right)\left(\mathbf{z}_{i}-\hat{\mu}_{k}\right)^{T}}{\sum_{i=1}^{N} \hat{\gamma}_{i k}}, \\
		\end{aligned}
	}\end{equation}
	where $\hat{\gamma}_{i}$ stands for the membership prediction, and $\hat{\phi}_{k}$, $\hat{\mu}_{k}$, $\hat{\mathbf{\Sigma}}_{k}$ are the mixture probability, mean and covariance for component $ k $ in GMM respectively, $\forall 1 \leq k \leq K$. Hence, the sample energy can be inferred as:
	\begin{equation}\label{eq_Energy}\resizebox{0.83\hsize}{!}{$
		E(\mathbf{z})=-\log \left(\sum_{k=1}^{K} \hat{\phi}_{k} \frac{\exp \left(-\frac{1}{2}\left(\mathbf{z}-\hat{\mu}_{k}\right)^{T} \hat{\mathbf{\Sigma}}_{k}^{-1}\left(\mathbf{z}-\hat{\mu}_{k}\right)\right)}{\sqrt{\left|2 \pi \hat{\mathbf{\Sigma}}_{k}\right|}}\right)$
	}\end{equation}
	where $ |\cdot| $ denotes the determinant of a matrix.

\subsection{Training Strategy}

The objective function of SOM-DAGMM is obtained as:
\begin{equation}\label{eq_obj_func}\resizebox{0.91\hsize}{!}{$
	J\left(\theta_{e}, \theta_{d}, \theta_{m}\right)=\frac{1}{N} \sum_{i=1}^{N} L\left(\mathbf{x}_{i}, \mathbf{x}_{i}^{\prime}\right)+\frac{\lambda_{1}}{N} \sum_{i=1}^{N} E\left(\mathbf{z}_{i}\right)+\lambda_{2} P(\hat{\mathbf{\Sigma}})$
}\end{equation}
where $L\left(\mathbf{x}_{i}, \mathbf{x}_{i}^{\prime}\right)$ is the reconstruction error of the autoencoder, $E\left(\mathbf{z}_{i}\right)$ indicates the sample energy and $P(\hat{\mathbf{\Sigma}})$ denotes a penalty term. \par

\begin{figure}[!tb]
	\centering
	\includegraphics[width=\columnwidth,height=2.52in]{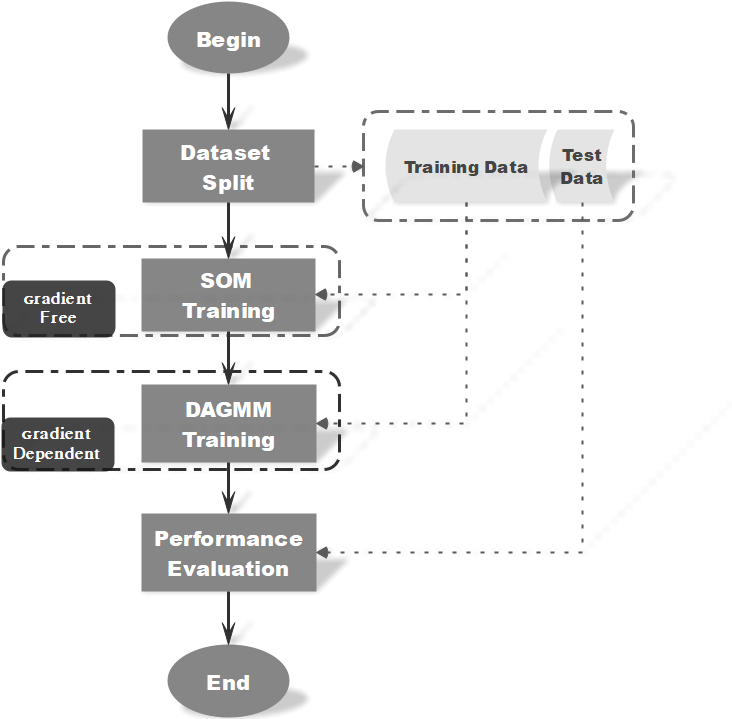}
	\caption{Training of Self-Organizing Map assisted Deep Autoencoding Gaussian Mixture Model.}
	\label{Fig_overview_training}
\end{figure}

Then, the training of the deep autoencoder and the multi-layer neural network is gradient dependent, while that of SOM is gradient-free and follows the competitive learning rule~\cite{kohonen1982self}. The training of the proposed SOM-DAGMM is divided into the SOM and the DAGMM sub-process, which is shown in Fig.~\ref{Fig_overview_training}. First, the dataset is split into two sets, including training and test data. Training set does not require labels given the unsupervised model, while those belonging to the test set are kept for performance evaluation.  \par

With the training data, SOM is subsequently trained following the competitive learning rule. Thereafter, the SOM-generated low-dimensional representation $\mathbf{z}_{s}$ is obtained. It is worth noting that the SOM together with its output, i.e. SOM-generated $\mathbf{z}_{s}$, remains fixed during the subsequent DAGMM training. As aforementioned, DAGMM gets its two components jointly trained, which corresponds to the joint optimization of the loss function defined in~\eqref{eq_obj_func}. Finally, performance of the obtained model is evaluated. Related details are given in Section~\ref{sec_4}. \par

\section{Experiments}\label{sec_4}
\subsection {Experimental Design}
In this section, extensive experiments have been conducted, and comparison has been made between the proposed SOM-DAGMM and the state-of-the-art DAGMM. We detail the \textit{Experiments on SOM Parameters} in \textit{Section~\ref{4_2}} and two scenarios that correspond to \textit{Section~\ref{4_3} Ideal Experiments assuming No Mixture in Training Data} and \textit{Section~\ref{4_4} Practical Unsupervised Experiments}. The ideal experiments assume that the data has been pre-processed to obtain the pure ``attack" instances and the two models are trained with these pure ``attack" instances. The second scenario involves the use of normal mixed data which have not been preprocessed which means it is more practical and valuable albeit more challenging. All the experiments are carried out ten times with a different fixed random seed each time. \par

\subsubsection {Dataset}
Two datasets including NSL-KDD~\cite{tavallaee2009detailed} and CSE-CIC-IDS2018~\cite{sharafaldin2018toward} are used in our experiments to evaluate the effectiveness of the proposed SOM-DAGMM. Table~\ref{tab_stat_datasets} shows the statistics of the two datasets. \par

KDDCUP99~ is the most well-known and widely-used dataset for the evaluation of intrusion detection and other anomaly detection approaches or systems. However, more and more defects, e.g., redundant records and unreasonable record proportions of difficulty levels have been reported in recent studies. By solving these inherent problems, NSL-KDD gives more objective and accurate evaluation. Records of NSL-KDD contain 41 features (34 continuous and 7 categorical) and are labeled as either normal or attack, with exactly one specific attack type. In pre-processing, continuous features are normalized with Min-Max Scaling, and categorical ones are converted to one-hot encoding representations with input dimension of 122. In NSL-KDD, all the involved attacks fall into the following four categories: \textbf{DoS attack}, \textbf{R2L}, \textbf{U2R}, and \textbf{Probing}. \par

As suggested in the paper proposing DAGMM~\cite{zong2018deep}, the majority group of KDDCUP99 that shares the same features with NSL-KDD, i.e. ``attack" instances, are adopted to train the model, whereas ``normal" instances constitute the minority group, i.e. the anomalies in this task. We also adopt the same approach as~\cite{zong2018deep, zhai2016deep} to evaluate the models. We take 50\% of the ``attack" instances through random sampling to train the models and evaluate them with the remaining 50\% and all the ``normal" instances. 

Furthermore, experimental studies adopt CSE-CIC-IDS2018 for making full comparison between the proposed SOM-DAGMM and the baseline algorithm, i.e. DAGMM. More insights into SOM-DAGMM are expected to be obtained as well. As a collaborative project between the Communications Security Establishment (CSE) \& the Canadian Institute for Cybersecurity (CIC), CSE-CIC-IDS2018 considers seven different attack scenarios, namely Brute-force, Heartbleed, Botnet, DoS, DDoS, Web attacks and infiltration.\par
\begin{table}[tb]
    \centering
    \caption{Statistics of Datasets}
    \begin{threeparttable}
    \resizebox{\hsize}{!}{
    \begin{tabular}{cccc}
        \toprule
        \textbf{Dataset} & \textbf{\# Dimensions} & \textbf{\# Instances} & \textbf{Anomaly Instance Ratio} \\
        \midrule
        NSL-KDD & {122}   & {4898431} & {46.54}\% \\
        CSE-CIC-IDS2018$^{\ast}$ & {115}   & {500000} & {26.28}\% \\
        \bottomrule
    \end{tabular}}%
    \begin{tablenotes}
            \item[$ \ast $] A 5\% subset is adopted for experimental assessments.
    \end{tablenotes}
    \end{threeparttable}
    \label{tab_stat_datasets}%
\end{table}%

\subsubsection{Setting on DAGMM}
For fair comparison, the hyper-parameters of both the DAGMM and those involved in our proposed SOM-DAGMM are set to be the same, which have been finely tuned by~\cite{zong2018deep}. The settings are summarized in Table~\ref{setting_DAGMM}. \par

\begin{table}[t!]
    \centering
    \caption{Default setting of DAGMM~\cite{zong2018deep}}
    \label{setting_DAGMM}
    \begin{tabular}{@{}p{3.6cm}<{\centering}p{3.6cm}<{\centering}@{}}
        \toprule
        \textbf{Notion} & \textbf{Value} \\ \midrule
        $\eta$ (Learning Rate) & 0.0001 \\ 
        $N$ (Batch Size)    & 1024 \\ 
        $\lambda_1$ (refer to Equation~\eqref{eq_obj_func}) & 0.1 \\
        $\lambda_2$ (refer to Equation~\eqref{eq_obj_func}) & 0.005 \\
        \bottomrule
    \end{tabular}
\end{table}

\subsection{Experiments on SOM Parameters}\label{4_2}

\begin{figure}[t!]
	\centering
	\includegraphics[width=\columnwidth]{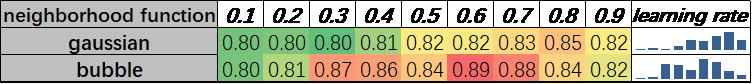}
	\caption{Performance of SOM-DAGMM on NSL-KDD over learning rate and neighborhood function.}
	\label{Fig_para_exp}
\end{figure}

\begin{table}[t!]
    \centering
    \caption{Default setting of SOM~\cite{vettigli2019minisom}}
    \label{setting_SOM}
    \begin{tabular}{ccc}
    \toprule
    \textbf{Hyper-Parameter} & \textbf{NSL-KDD} & \textbf{CSE-CIC-IDS2018} \\
    \midrule
    \textbf{learning rate} & 0.6   & 0.8 \\
    \textbf{neighborhood function} & bubble & bubble \\
    \bottomrule
    \end{tabular}
\end{table}
In the implementation, MiniSom is employed for constructing the SOM-DAGMM~\cite{vettigli2019minisom}. Two key hyper-parameters are tuned through grid search and Table~\ref{setting_SOM} shows optimal results corresponding to the two datasets. Furthermore, Fig.~\ref{Fig_para_exp} details the tuning experiment for NSL-KDD for a better understanding. Here, we consider \textbf{F1 Score} as the metric for performance evaluation.\par 

\subsection{Ideal Experiments Assuming No Mixture in Training Data}\label{4_3}

In this scenario, the two models involved are trained with pure ``attack" instances. Four metrics are adopted for measuring the model performance. They are \textbf{Accuracy}, \textbf{Precision}, \textbf{Recall} and \textbf{F1-Score}. 

All the experiments are independently run for ten times. Table~\ref{tab:pre-filtered} gives the corresponding average (AVG) and standard deviation (STDEV) values with the AVG value being listed before the standard deviation which is in parentheses. 
Given the results, the following conclusions can be drawn:
\begin{itemize}
    \item In both datasets, {SOM-DAGMM} performs better than DAGMM in all the four metrics.
    \item Moreover, {SOM-DAGMM} achieves a more stable performance as manifested in the STDEV values.
    \item These improvements, including better overall performance and stability, prove the contribution of the input topology brought about by SOM, together with the well-designed model architecture and training.
\end{itemize} 

\begin{table}[t!]
  \centering
  \caption{Performance of model trained with only normal data}
  \resizebox{\hsize}{!}{
    \begin{tabular}{ccccc}
    \toprule
    \textbf{Dataset} & \multicolumn{2}{c}{\textbf{NSL-KDD}} & \multicolumn{2}{c}{\textbf{CSE-CIC-IDS2018}} \\
\cmidrule{2-5}    \textbf{Algorithm} & \textbf{DAGMM} & \textbf{SOM-DAGMM} & \textbf{DAGMM} & \textbf{SOM-DAGMM} \\
    \midrule
    \textbf{Accuracy} & 0.71(0.12) & \textbf{0.85(0.02)} & 0.83(0.02) & \textbf{0.89(0.01)} \\
    \textbf{Precision} & 0.82(0.08) & \textbf{0.91(0.01)} & 0.93(0.01) & \textbf{0.95(0.00)} \\
    \textbf{Recall}& 0.73(0.11) & \textbf{0.86(0.02)} & 0.87(0.01) & \textbf{0.91(0.01)} \\
    \textbf{F1 Score}& 0.77(0.10) & \textbf{0.89(0.01)} & 0.90(0.01) & \textbf{0.93(0.01)} \\
    \bottomrule
    \end{tabular}}
  \label{tab:pre-filtered}%
\end{table}%

\subsection {Practical Unsupervised Experiments}\label{4_4}
\begin{table*}[t!]
\centering
\caption{Performance of model trained with mixed data}
  \begin{tabular}{ccccccc}
  \toprule
  \textbf{Anomaly Ratio} & \multicolumn{2}{c}{\textbf{1\%}} & \multicolumn{2}{c}{\textbf{5\%}} & \multicolumn{2}{c}{\textbf{10\%}} \\
\cmidrule{2-7}    \textbf{Algorithm} & \textbf{DAGMM} & \textbf{SOM-DAGMM} & \textbf{DAGMM} & \textbf{SOM-DAGMM} & \textbf{DAGMM} & \textbf{SOM-DAGMM} \\
  \midrule
  \textbf{Accuracy} & 0.71(0.11) & \textbf{0.85(0.02)} & 0.69(0.09) & \textbf{0.83(0.02)} & 0.65(0.10) & \textbf{0.78(0.04)} \\
  \textbf{Precision} & 0.81(0.08) & \textbf{0.90(0.01)} & 0.79(0.07) & \textbf{0.89(0.01)} & 0.75(0.08) & \textbf{0.85(0.02)} \\
  \textbf{Recall} & 0.72(0.11) & \textbf{0.86(0.02)} & 0.71(0.09) & \textbf{0.84(0.02)} & 0.67(0.10) & \textbf{0.80(0.03)} \\
  \textbf{F1 Score} & 0.76(0.10) & \textbf{0.88(0.01)} & 0.75(0.08) & \textbf{0.86(0.01)} & 0.71(0.09) & \textbf{0.82(0.03)} \\
  \bottomrule
  \end{tabular}
\label{tab:practical}%
\end{table*}%

\begin{figure}[t!]
	\centering
	\includegraphics[width=\columnwidth, height=1.36in]{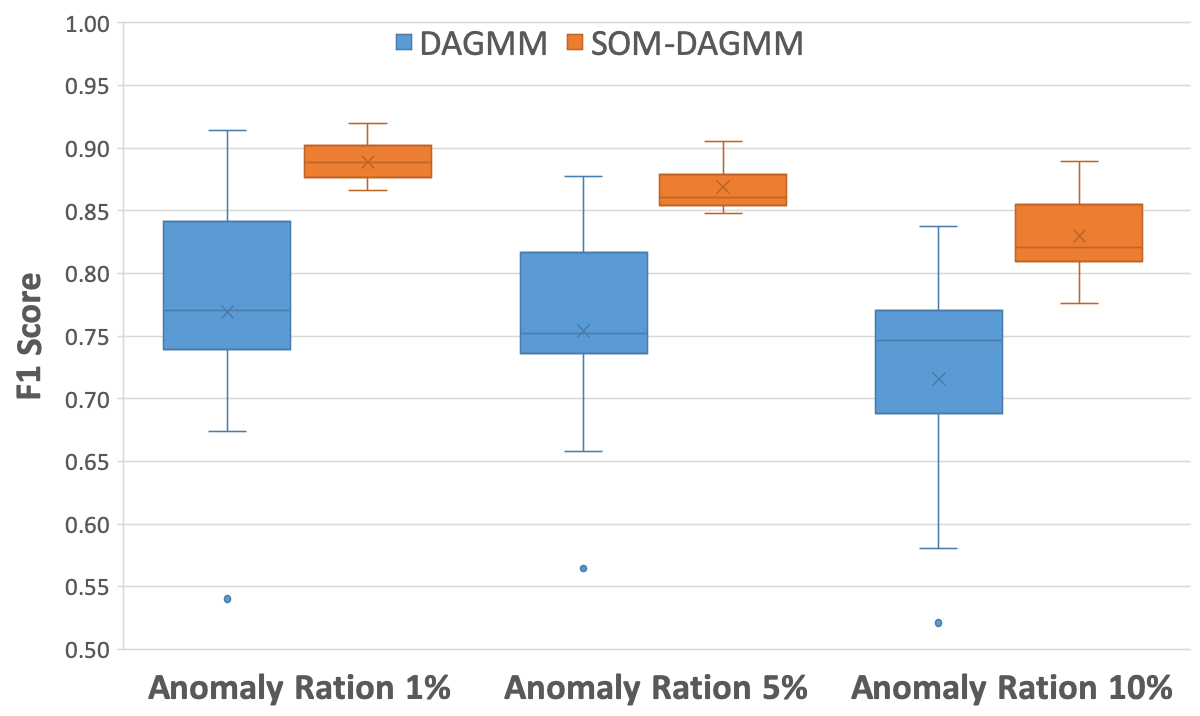}
	\caption{Whisker Plot: F1 Score performance on NSL-KDD.}
	\label{F1_Box_Plot}
\end{figure}

Most of the settings, including performance measures, the adopted random seeds and all the hyper-parameters, are the same as the previous subsection except for how the training data is being dealt with. 
However, the assumption that pre-filtering gets rid of all ``non-attack" instances is too ideal for most intrusion detection techniques to hold. Thus, it is necessary to evaluate the models on training data that are mixed with unexpected instances since pure training data is far-fetched. \par 

\begin{figure}[t]
	\centering
	\includegraphics[width=\columnwidth, height=1.25in]{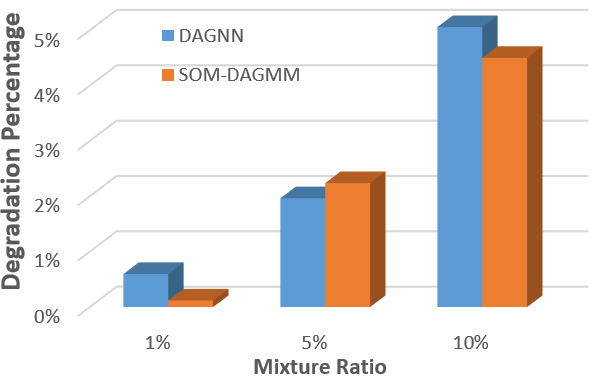}
	\caption{F1 Score Degradation on NSL-KDD.}
	\label{Fig_Degradation}
\end{figure}

All the experiments are independently conducted for ten times. Their AVG and STDEV values are listed in Table~\ref{tab:practical}. In addition, F1 Scores are illustrated as the whisker plot in Fig.~\ref{F1_Box_Plot}. Based on the results, similar conclusions can be drawn:
\begin{itemize}
    \item 
    Similar to Section~\ref{4_3}, 
    {SOM-DAGMM} outperforms DAGMM in all the metrics for both datasets. 
    \item Moreover, {SOM-DAGMM} achieves more stable performance when considering STDEV values. 
    \item In contrast to the previous scenario where only ``attack" instances are used for training, performance deterioration of both SOM-DAGMM and DAGMM is observed in all metrics as expected. As shown in Fig.~\ref{Fig_Degradation}, SOM-DAGMM does not achieve significant improvement in robustness and this will be addressed in our future studies. 
\end{itemize} 

\section{Conclusion}\label{sec_5}

To preserve the input space topology that is inadequately handled by the deep autoencoder in DAGMM, we propose a self-organizing map assisted deep autoencoding Gaussian mixture model (SOM-DAGMM) to achieve performance improvement with the help of SOM. 
The single most striking observation which emerges from the experiments performed on the datasets is the superiority and the stability of SOM-DAGMM in all the standard metrics. Moreover, in a more practical, valuable and challenging scenario where mixed training data are used, SOM-DAGMM is able to rise to the challenge and achieve good performance results. \par

Federated learning has recently been applied in secure machine learning where the server only carries out parameter aggregation and keeps private data locally stored so that privacy is properly guaranteed~\cite{chen2019communication, chen2019federated}. For future work, we will develop a new federated learning assisted SOM-DAGMM to seek further performance improvement through increasing data availability while maintaining data privacy. \par

\bibliographystyle{IEEEtran}
\bibliography{IEEEabrv,Bib0118}

\end{document}